# The Relevance of Proofs of the Rationality of Probability Theory to Automated Reasoning and Cognitive Models

**Ernest Davis, Computer Science Department, New York University**

## Abstract

*A number of well-known theorems, such as Cox's theorem and de Finetti's theorem. prove that any model of reasoning with uncertain information that satisfies specified conditions of "rationality" must satisfy the axioms of probability theory. I argue here that these theorems do not in themselves demonstrate that probabilistic models are in fact suitable for any specific task in automated reasoning or plausible for cognitive models. First, the theorems only establish that there exists some probabilistic model; they do not establish that there exists a useful probabilistic model, i.e. one with a tractably small number of numerical parameters and a large number of independence assumptions. Second, there are in general many different probabilistic models for a given situation, many of which may be far more irrational, in the usual sense of the term, than a model that violates the axioms of probability theory. I illustrate this second point with an extended examples of two tasks of induction, of a similar structure, where the reasonable probabilistic models are very different.*

Advocates of probabilistic methods in artificial intelligence (AI) and cognitive modeling have often claimed that the only rational approach to representing and reasoning with uncertain knowledge to use models based on the standard theory of probability; and that the only rational approach to making decisions with uncertain knowledge is to use decision theory and the principle of maximum expected utility. Moreover it is stated that this claim is in fact a mathematical theorem with well-known proofs such as those of Cox (1946) (1961), de Finetti (see for example the discussion in (Russell & Norvig, 2009) ), Savage (1954) and so on. For example, Jacobs and Kruschke (2011) write, "Probability theory does not provide just any calculus for representing and manipulating uncertain information, it provides an *optimal* calculus" (emphasis theirs). Part of this claim, sometimes made explicitly, is that a reasoner can, whenever it needs to, assign a probability to any given meaningful proposition, and a conditional probability to every pair of propositions, in a way that is, over all, consistent with the axioms of probability theory.

The usefulness in practice of probabilistic models for at least the current generation of AI systems is indisputable. There is also much evidence that probabilistic models are often useful for cognitive models, though I have argued elsewhere (Marcus & Davis, to appear) that some claims that have been made are overstated. I do not, in this paper, discussing the empirical evidence for either of these points; I discuss only the relevance of the *mathematical proofs.* In particular I address two questions. First, do the mathematical proofs add any significant support to the belief that probabilistic models will be useful for AI and for cognitive models? My answer is, only very moderately. The second question is, do the mathematical proofs indicate that researchers should not spend their time on non-probabilistic models, as inherently suboptimal? My answer is, not at all.

First, it is important to have a clear idea of what these theorems actually state. Cox's theorem proves that if a reasoner assigns numerical probabilities to propositions and updates these on the appearance of new evidence, and these assignments and updatings satisfy certain specified canons of rationality[1], then the assignment satisfies the axioms of probability theory. De Finetti's theorem proves that if a reasoner accepts hypothetical bets before and after receiving information, and this system of bets satisfies canons of rationality, then the bets correspond to a judgment of probabilities that satisfies the axioms of probability theory. Specifically, if the system of bets violates the axioms of probability theory, then it is possible to create a "Dutch book" against the reasoner, a set of bets, each of which individually the reasoner would accept, but which in combination are guaranteed to lose money. Savage's theorem proves that if a reasoner is given a collection of choices between lotteries, and his choices satisfies canons of rationality, then there exists an assignment of probabilities satisfying the axioms of probability theory and an assignment of utilities to the outcomes, such that the reasoner is always choosing the maximum expected utility outcome. In this paper, I will not address Savage's theorem, though similar considerations apply, and for convenience I will use the phrase "Cox's theorem" to refer generically to Cox's theorem, de Finetti's and other similar justifications of the axioms of probability as the only rational basis of a general theory of reasoning with uncertainty.

What these and similar theorems accomplish is to offer elegant arguments that the axiomatic theory of probability theory, which was developed in the context of a sample space interpretation, and the calculation of expected utility, which was developed in the context of gambling for money, can be reasonably applied in the much broader setting of uncertainty of any kind and preferences of any kind.

The theorems also plausibly support the following statements. In automated reasoning, if there exists a solidly grounded effective probabilistic model for a domain, then you will generally do better applying the standard theorems of probability theory to this model than using some other mathematical manipulation over the numbers in the model. Alternatively, if some other mathematical manipulation actually does yield more useful answers --- this is, after all, an empirical question --- then that is a fact that calls for some explanation; there is presumably a bug in the probabilistic model, which can perhaps be characterized. For cognitive models, if the likelihoods that a human reasoner assigns to various propositions can be reliably established, and these likelihoods violate the theory of probability, then there is justification for calling his reasoning processes irrational. If they conform to the theory of probability, then they can be taken as rational at least in that respect.

There is a large, acrimonious literature on the reasonableness of the various canons of rationality that these arguments presume; but for argument's sake, let us here accept the premises of these proofs, and therefore let us accept that the conclusions of the proof are valid. Even so, however, these proofs say almost nothing about what a model of reasoning or action for any given task should or should not look like, because the axioms of probability are very weak constraints. For example, the axioms give almost no constraint on belief update. If you have a conclusion X and a collection of 6 pieces of evidence A...F, then *any* assignment of values between 0 and 1 to the 64 different conditional probabilities $P(X)$, $P(X|A)$, $P(X|B)$ ... $P(X|F)$, $P(X|A,B)$, $P(X|A,C)$ ... $P(X|A,B,C,D,E,F)$ is consistent with the axioms of probability theory. All that the axioms of probability prohibit are things like two synonymous statements having different probabilities, or the conjunction fallacy $P(A,B) > P(A)$, both of which, of course, have been demonstrated by Kahneman and Tversky (1982) and others to occur in human evaluations of likelihood.

In particular, I note a number of serious limitations on the scope of these theorems and therefore on their relevance to AI and cognitive modeling.

---

[1] The word "optimal" in the above quotation from Jacobs and Kruschke is poorly chosen. In each of these proofs, a behavior or assignment that violates the conditions is irrational, rather than suboptimal.

First, all that the proofs establish is that there exists *some* probabilistic model. In a situation where one is considering *k* different propositions, a general probabilistic model can have as many as $2^k$-1 degrees of freedom (logically independent numerical parameters); in fact, mathematically almost all models do have $2^k$-1 degrees of freedom. If *k* is of any substantial size, such a model is entirely useless. A *useful* probabilistic model is one that has a much smaller number of degrees of freedom, typically a small constant times *k* or at most times $k^2$. This is usually attained by positing a large number of independence assumptions. However, the proofs give no reason whatever to expect that there exists a probabilistic model of the judgments of likelihood that is useful in this sense.

In fact non-probabilistic models can often be as effective that as probabilistic model and simpler. Consider the following example. Suppose that a job interview involves a placement exam consisting of 10 questions; that the exam is graded pass/fail, with a passing mark of 5 or higher; and that passing the exam is largely determinative of a job offer. 90% of applicants who fail are rejected and 90% who pass are accepted, the additional variance being uncorrelated with the scores on the exam questions. Suppose that Alice, Bob, Carol, and Doug are developing automated systems to predict the hiring decision from the test answers.

Alice is not a probabilist. She writes a system that applies the rule that applicants who score 5 or higher are offered a job, and notes that her system predicts the correct result 90% of the time.

Bob believes in Bayesian networks. He produces the network shown in figure 1. The associated table of conditional probabilities has 1024 separate numerical parameters, which he will use machine learning techniques to train. This will take a data set of about a million elements to do with any significant accuracy. The Bayesian network also expresses the statement that each of the questions is independent of all the rest; this will take a comparable number of data items to check; if false, a more complex network will be needed. Fortunately, Bob is also a believer in Big Data, so he is not fazed.

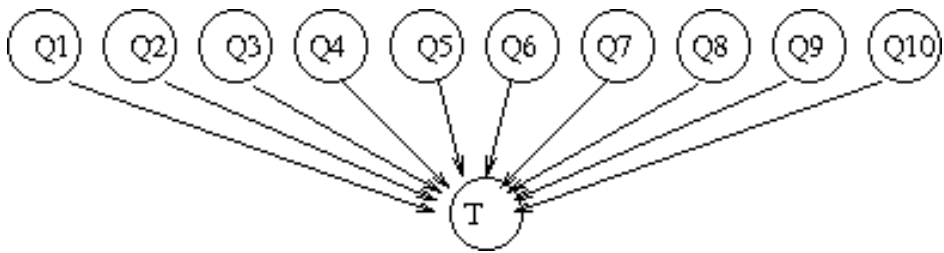

Figure 1: Bayesian network

Carol believes in Bayesian networks with hidden variables. She produces the network shown in figure 2. This has only 84 numerical parameters, and thus requires much less data to train. However, more sophisticated ML techniques are needed to train it, particularly as Carol wishes to automatically learn the significance of the hidden variables rather than hand-code them (the labels under the hidden nodes in figure 2 are purely for the convenience of the reader of this paper).

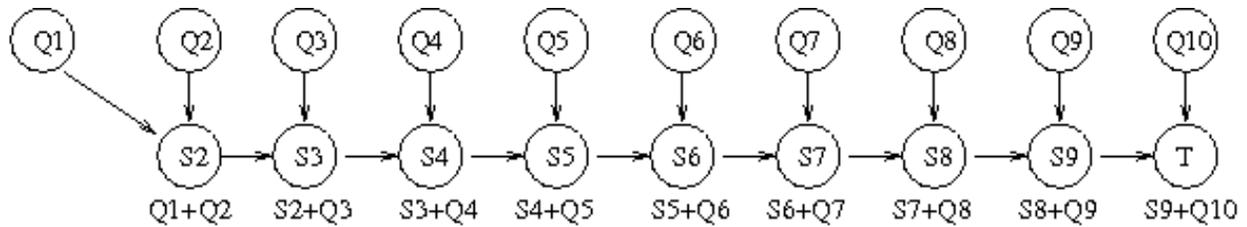

Figure 2: Bayesian network with hidden variables

Doug takes a broader view of probabilistic models. He produces the following probabilistic model

S = [(Q1 + Q2 + Q3 + Q4 + Q5 + Q6 + Q7 + Q8 + Q9 + Q10) ≥ 5]

With probability 0.9, T=S; else T=~S.

This avoids the problems of Bob's and Carol's models. However, it obviously has essentially zero added value as compared to Alice's.

Second, given a choice between two specific models, one of which is probabilistic and the other of which is not, the theorems give no warrant whatever for supposing that the probabilistic model will give better results, in an AI setting, or a more accurate cognitive model, in a psychological setting, even if one supposes in the latter case that people are on the whole "rational". It may be an empirical fact that probabilistic models have in fact worked well in both these setting, but that fact has essentially no relation to these theorems. Probabilistic models can be useless in the AI setting and remote from cognitive reality in the cognitive setting; and these theorems are just as happy with a useless or false model as with a valid one. The point is too obvious to require elaboration; but the false presumption is nonetheless often made.

Third, there is very little reason to believe that "the prior subjective probability of proposition Φ" is in general a stable or well-defined cognitive entity. Subjects in psychological experiments tend to be cooperative, and if you ask them to give you a number between 0 and 1 characterizing their belief in Φ, they will happily give you a number. However, that is a number that has been elicited by the experimental procedure. It may well have only a remote relation to any characteristic of their mental state before being asked the question; and alternative procedures will get you different numbers.

## Examples

Let me illustrate these points with an extended hypothetical example in cognitive modeling. Consider a psychologist who is studying the inductions of universal generalizations "All X's are Y."  Imagine that he carries out the following experiment:

**Experiment 1:** The target generalization is "All Roman Catholic priests are male"; we will call this proposition Φ. (As of the time of writing, this is a true statement.)  For this purpose the experimenter selects subjects who, because of their milieu or age, are unaware of this fact. He informs the subjects that there are about 400,000 Roman Catholic priests in the world. He then shows the subjects three photographs of Roman Catholic priests, all male, in succession. Before showing any photographs, and after each photograph, he asks them what likelihood they assign to the statement "All Roman Catholic priests are male."

A Bayesian theorist might reasonably propose the following model for the cognitive process involved here:

**Model A:** The subject considers the random variable $F$, defined as "the fraction of priests who are male"; thus $\Phi$ is the event $F=1$. Let $M_k$ be the event that a random sample of $k$ priests are all male.
The subject has a prior distribution over $F$ that we will define below. The subject assumes that the photos being shown are a random sample from the space of all priests, or at least a sample that is independent of gender. After seeing $k$ photos, he compute the posterior conditional probability of the event $F=1$ given $M_k$. He reports that posterior probability as his current judgment of the likelihood.

We assume that the subjects, after seeing a few photographs of male priests, assign some significantly large value to the likelihood of $\Phi$. Then the prior probability of $\Phi$ cannot be very small. For instance, if $F$ is uniformly distributed over [0,1], then after the subject has seen 1000 photos, the posterior probability that all 400,000 priests are male is 0.0025, (After all, if only 399,000 of the 400,000 are male, the probability of $M_{1000}$ is still almost 0.78.) On the other hand, the subject presumably does not start with the presumption that necessarily $F=1$ or $F=0$; it could certainly be a mixture. Finally, it is certainly reasonable to suppose that the subject considers males and females symmetrically for this purpose.

These considerations suggest the following prior distribution for $F$. The distinguished values $F=1$ and $F=0$ have some particular probability $p$ which is not very small. The remaining $1-p$ worth of probability is uniformly distributed over (0,1); that is, for $0 < x < 1$, the probability density of $F=x$ is $1-2p$.

Given this model, it is straightforward to show, using Bayes' law, that the conditional probability Prob($\Phi | M_k$) for $k=0$ is $p$, and for $k \geq 1$ is

$$\frac{(k+1)p}{1+(k-1)p} + \frac{(1-2p)}{(n+1)[1+(k-1)p]}$$

where $n$=400,000 is the size of the population. (The first term corresponds to the probability that $F=1$; the second is the probability that $\Phi$ is true even if $F < 1$.) For example for $p$=0.1, $k$=11, we have Prob($\Phi | M_k$) = 0.6. The induction seems slow --- people's ability, or willingness, to make strong generalizations on the basis of very small numbers of examples is well-known to be hard to explain --- but broadly speaking the model is doing the right kind of thing.

Consider, however, the following alternate prior distribution:

**Model B**: Each priest is randomly and independently either male or female, with probability 1/2. The prior distribution of $F$ is therefore the binomial distribution B(0.5, 400000).

Given this prior distribution, the posterior probability Prob($\Phi | M_k$) = $2^{-(400000-k)}$. In this model, when you find out that one priest is male, the only information that gives you is the sex of that one priest; since the other priests are independent, they still each have a 1/2 chance of being female. After you have seen k male priests, the probability that the remaining 400000-k priests are all male is therefore = $2^{-(400000-k)}$.

Model B has a certain elegance, but it is obviously useless for induction; the only way to induce a generalization is to see every single instance. It is obviously absurd and not worth considering. Except that, for experiment 2, it is the correct model.

**Experiment 2:** Identical to experiment 1, except that the hypothesis now under discussion is "All of the babies born in Brazil last year were male." The experimenter shows a series of photos of male Brazilian babies.

For experiment 2, clearly model B is appropriate, or at least much more nearly so than model A. Estimating the annual births in Brazil at about 5 million yields a prior probability of $2^{-5000000}$ for $\Phi$; that does not seem unreasonable.

It is interesting to consider what a reasonable subject would conclude as the experimenter shows him one photo after another of a male Brazilian baby. It seems safe to conjecture that the subject will fairly soon conclude that these are not a random sampling of Brazilian babies and will stick with that conclusion. If the experimenter insists that they are, then the most reasonable conclusion is that the experimenter is either lying, mistaken, or insane. This possibility can, of course, be incorporated in our model by using a mixed model in which, with probability *p*, the sample is a random one, and with probability *1-p* it was deliberately selected to be all males. The posterior probability of the hypothesis that it is a random sample then rapidly goes to 0.

Suppose, however, that the subject for some reason has absolute faith in the experimenter's statement that this is indeed a random sample. That's actually too hard to believe; but the subject might be willing to entertain the statement as a hypothesis: "Suppose for argument's sake that this is a random sample; then what would you conclude?" In that case, my guess is that it will still take quite a few photographs before the subject starts to give serious consideration to the question, "What in the name of God is going on in Brazil?", because there really is no reasonable explanation of how this could happen. Moreover, even once the subject has decided that something very strange is happening in Brazil, it may still require many more photographs before he assigns a large probability to the event that *every* baby in Brazil -- in the cities, in the country, in the slums, in the rainforest --- was born male.

There are several points here. First, obviously, there are here two very simple, standard probabilistic models giving two drastically different predictions for two ostensibly similar situations. Cox's proof gives no guidance as which model should be used in which experiment. Principles such as maximum entropy are no more helpful. If one considers a model in which the sex of the *k*th person is a random variable $X_k$, then the maximum entropy assumption yields model B. Perhaps one could contrive a system of random variables over which the maximum entropy assumption would yield model A, or something similar. But whether or not, there does not seems to be any way to use maximum entropy arguments to choose model A for the first experiment and model B for the second.

Second: It seems clear that, in the ordinary sense of "rationality", a subject who used Model A for Experiment 2 or Model B for Experiment 1 would be far more irrational than the subjects who committed the conjunction fallacy in the famous "feminist bank teller" experiment of Kahneman and Tversky.

Third: The problem of how world knowledge is used to choose the proper model in a given situation is an important one, on which, to the best of my knowledge, little has been done in either the AI, the cognitive, or the philosophical literature.

Fourth: The arguments above are in the wrong direction. In developing the models above, I did not actually consider what probabilistic models were appropriate and derive their consequences for the subjects' answers; I considered what subjects would be likely to answer in the experiment and designed the models to fit them, extending them to be theoretically ugly mixed models when that was needed. This is not, I think, merely a rhetorical trick on my part as author here; my guess is that anyone developing a probabilistic model for these kinds of situations would do likewise. If that is correct, what that suggests rather strongly is that, in the minds of the theorists, the responses are epistemically primary and that the probabilistic models are derived from them. That in turn suggests, though not as strongly, that the responses are what is cognitive real here in the minds of the subjects, and that the probabilistic models are theoretical fictions.

Fifth: Of course, one can "explain" the choice between the two models in an overarching Bayesian model. Simply construct a mixed model in which, with probability *q*, model A applies and with probability *1-q* model B applies; and then given a sample, the posterior probability of the wrong model rapidly goes to zero. But this is not very satisfying. For one thing, this mixed model is clearly constructed post hoc. For another thing it has the undesirable characteristic that, the less you know, the more complicated the model. The point of subjective probability theory is to give an account of reasoning with limited information. But here, you get a simple stochastic model only when you have a lot of meta-level information; namely, you know which model applies. When you really know little, then you need a horrific mixed model, which subsumes a large number of specific models, and assigns a largely arbitrary prior probability to each of them.

Though widely different, the two above models do have the feature that the posterior probability of Φ is monotonically non-decreasing. Indeed, any model in which the sample is drawn randomly following a given distribution, and the probability of the data given the hypothesis is computed as the probability of a sample with this number of males given the distribution will likewise be monotonically non-decreasing. Suppose however that we run experiment 1, and we encounter the following pattern of subject responses:

| After     | Subject |
|-----------|---------|
| 0 photo   | 0.1     |
| 1         | 0.9     |
| 2         | 0.9     |
| 3 or more | 0.1     |

Table 2: Hypothetical data

This subject's responses seem strange. However, even he is not necessarily irrational in the sense of Cox's theorem, just idiosyncratic. His answers can be justified n terms of the following probabilistic model. Let Φ be the proposition "All priests are male". For *k*=0,1,2 let $\lambda_k$ be the proposition, "The experimenter will show me exactly *k* photos," and let $\lambda_{>2}$ be the proposition "The experimenter will show me more than 2 photos." Note that after seeing 1 photo, the subject can rule out $\lambda_0$, but the other options are possible, and similarly at the other values of *k*. The subjects' responses can then be '"explained" by positing the following priors and likelihoods and applying Bayes' law:

| Subject | P(Φ) | P($\lambda_0$\|Φ) | P($\lambda_0$\|~Φ) | P($\lambda\Phi_1$\|) | P($\lambda_1$\|~Φ) | P($\lambda_2$\|Φ) | P($\lambda_2$\|~Φ) | P($\lambda_{>2}$\|Φ) | P($\lambda_{>2}$\|~Φ) |
|---|---|---|---|---|---|---|---|---|---|
| 1 | 0.1 | 0 | 0 | 0 | 0 | 0 | 0.9877 | 1 | 0.0123 |
| 2 | 0.1 | 0 | 0.987 | 0 | 0 | 0.0987 | 0 | 0.0123 | 0.0123 |

Table 3: A probabilistic model for the data in table 2

It may be noted, incidentally, that positing that the subjects are taking the experimenters' intentions into account is perfectly kosher; exactly this is done, for example, Gweon, Tenenbaum, and Schultz (2010)

I am not claiming that the use of Bayesian models in the psychological literature is as arbitrary as table 3, though I have demonstrated elsewhere (Marcus & Davis, to appear) that it can certainly tend in this direction. The first point here is merely that Cox's theorem does not by any means exclude or deprecate this kind of model. The second point is that the probabilistic model in table 3 has exactly no actual explanatory value for the data in table 2. There is no advantage to using the probabilities in table 3 as a theory over simply using the numbers in table 2. Now, there are many possible choices for the numbers in table 3 that will match the data of table 2 — these particular values were chosen purely for the convenience of the authors — and one could probably come up with a more "principled" set of numbers by using considerations of maximum entropy or something similar. But those alternative probabilistic models would also offer no actual advantage over the raw data.